\documentclass[sigconf]{acmart}
\usepackage{tabularx}
\usepackage{makecell}
\usepackage{multirow}
\AtBeginDocument{%
  }

\copyrightyear{2026}
\acmYear{2026}
\setcopyright{cc}
\setcctype{by}
\acmConference[SIGIR '26]{Proceedings of the 49th International ACM SIGIR Conference on Research and Development in Information Retrieval}{July 20--24, 2026}{Melbourne, VIC, Australia}
\acmBooktitle{Proceedings of the 49th International ACM SIGIR Conference on Research and Development in Information Retrieval (SIGIR '26), July 20--24, 2026, Melbourne, VIC, Australia}
\acmDOI{10.1145/3805712.3808459}
\acmISBN{979-8-4007-2599-9/2026/07}

\begin{document}

\title{Probe-then-Plan: Environment-Aware Planning for Industrial E-commerce Search}


\author{Mengxiang Chen}
\authornote{The corresponding author.}
\email{chenmengxiang9@jd.com}
\affiliation{%
  \institution{JD.com}
  \city{Beijing}
  \country{China}
}

\author{Zhouwei Zhai}
\orcid{1234-5678-9012}
\email{zhaizhouwei1@jd.com}
\affiliation{%
  \institution{JD.com}
  \city{Beijing}
  \country{China}
}

\author{Jin Li}
\email{lijin.257@jd.com}
\affiliation{%
  \institution{JD.com}
  \city{Beijing}
  \country{China}
}

\renewcommand{\shortauthors}{Chen et al.}


\begin{abstract}
Modern e-commerce search is evolving from simple keyword matching to resolving complex user intents. While large language models (LLMs) offer powerful reasoning capabilities, existing LLM-based search paradigms suffer from a fundamental blindness-latency dilemma: query rewriting methods are agnostic to retrieval tool capabilities and real-time inventory states, resulting in invalid plans; conversely, deep search agent approaches initially plan without environment awareness, then rely on iterative tool calls and reflection to perceive and correct failures, leading to seconds of latency, incompatible with the sub-second budget for the planning module in industrial e-commerce search. To resolve this conflict, we propose Environment-Aware Search Planning (EASP), a novel paradigm that reformulates search planning as a dynamic reasoning process grounded in environmental reality. EASP introduces a Probe-then-Plan mechanism: a lightweight Retrieval Probe first exposes the retrieval snapshot, enabling the Planner to diagnose execution gaps and generate grounded search plans. Our methodology unfolds in three stages: (1) Offline Data Synthesis: The Teacher Agent synthesizes diverse, execution-validated plans by diagnosing the retrieval environment exposed by the Retrieval Probe. (2) Planner Training and Alignment: The Planner is initialized via supervised fine-tuning (SFT) on the offline dataset to internalize the Teacher's diagnostic capabilities, followed by alignment with business outcomes (conversion rate) through reinforcement learning. (3) Adaptive Online Serving: A complexity-aware routing mechanism selectively activates the planning pipeline only for complex queries, ensuring optimal resource allocation.
Extensive offline evaluations and online A/B testing on JD.com demonstrate that EASP significantly improves relevant recall and achieves substantial lifts in UCVR and GMV. EASP has been successfully deployed in JD.com's AI-Search system.
\end{abstract}



\maketitle

\begin{figure*}[t] 
  \centering
  \includegraphics[width=0.9\textwidth]{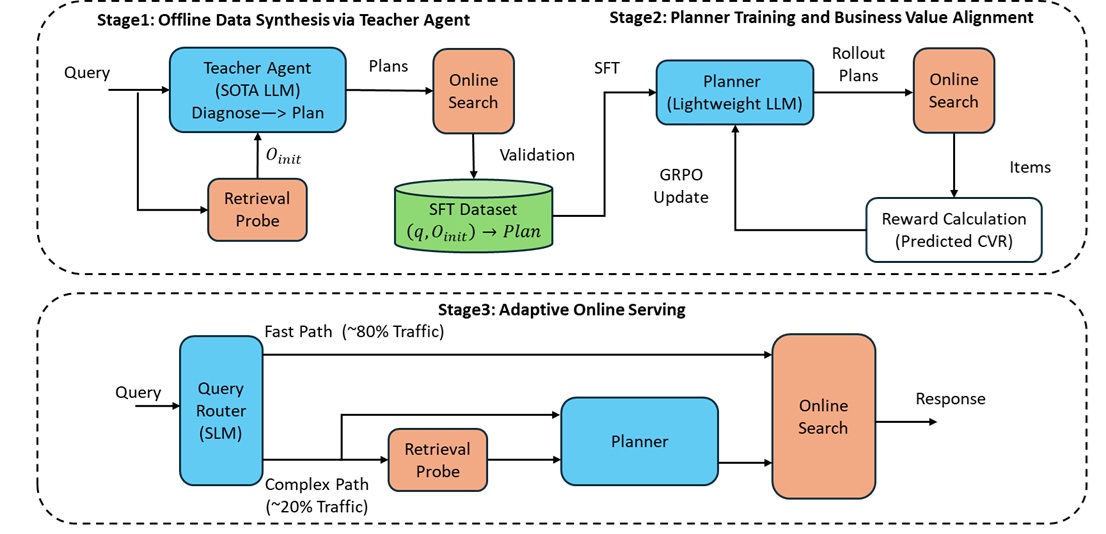} 
  
  \caption{The training and serving workflow of EASP}
  
  
  \label{fig:planner} 
\end{figure*}

\section{Introduction} \label{sec:introduction}

Modern e-commerce search is evolving from fulfilling explicit queries to resolving complex exploration scenarios with implicit constraints (e.g., ``bottoms match green shirts'', ``all-day comfort heels for the office''). Traditional term matching\cite{robertson2009probabilistic,aizawa2003information,deerwester1990indexing} and embedding-based retrieval\cite{karpukhin2020dense,Li2021EmbeddingbasedPR,Kekuda2024EmbeddingBR,He2023Que2EngageER} often fail to decompose such intents, causing recall deficiencies or semantic drift. While LLMs\cite{achiam2023gpt,touvron2023llama,guo2025deepseek} offer strong reasoning capabilities, existing solutions suffer from a fundamental blindness-latency dilemma: generative rewriting methods\cite{peng2024large,dai2024enhancing,nguyen2025minielm} operate blindly without perceiving initial query retrieval outcomes, producing redundant or infeasible queries; conversely, ReAct\cite{yao2022react}-style deep search agents\cite{jin2025search,qu2025tool,li2025websailor,hu2025step,tongyidr} first generate plans without environment awareness, then depend on iterative tool calls and reflection to perceive and correct failures, which incurs seconds of latency, making them impractical for real-time industrial deployment where planning must complete within hundreds of milliseconds.

Distinct from open-domain deep search tasks that necessitate multi-hop reasoning, complex e-commerce queries typically operate within a more bounded semantic space. We observe that a single diagnostic reflection on initial retrieval results is sufficient to formulate an effective search plan for most complex e-commerce queries. 

Motivated by this observation, we propose Environment-Aware Search Planning (EASP), a paradigm that reformulates query planning as a grounded, single-step reasoning process. EASP introduces a Probe-then-Plan mechanism: a lightweight Retrieval Probe first exposes a snapshot of the retrieval environment, enabling the Planner to diagnose execution gaps (e.g., entity drift, attribute misalignment) and generate grounded search plan in one go.
Online A/B testing on JD.com live traffic shows that EASP achieves statistically significant lifts in UCVR (+0.89\%, p<0.05) and GMV (+0.57\%, p<0.05), with end-to-end planner latency under 20ms at p75 (dominated by simple queries that bypass the planning pipeline) and under 700ms at p99 (cost of complex queries that trigger full planning) — well within industrial deployment constraints. Based on these results, EASP has been fully deployed in JD.com's AI-Search system.

In summary, our contributions are threefold: (1) an e-commerce search planning paradigm that grounds single-step planning in retrieval reality via Probe-then-Plan; (2) a training framework that enables a lightweight planner to acquire diagnostic capabilities from a teacher through supervised fine-tuning and align with business objectives via reinforcement learning; (3) a deployable industrial architecture that balances latency and effect through complexity-aware routing.

\section{Methodology}
\label{sec:methodology}

The overall architecture of our proposed framework is illustrated in Figure \ref{fig:planner}.

\subsection{Problem Formulation}
We formulate the search planning task as a decision-making process within a dynamic retrieval environment. Let $\mathcal{E}$ denote the e-commerce search engine environment, which contains the complete product inventory and retrieval logic.
Given a user query $q$, the true state of $\mathcal{E}$ is partially observable.

To initiate the planning process, a \textit{Retrieval Probe} is called to interact with $\mathcal{E}$, yielding an initial observation $\mathcal{O}_{\text{init}} = \text{Probe}(\mathcal{E}, q)$. This observation serves as a snapshot of the retrieval reality relative to $q$.
The planner functions as an agent policy $\pi_{\theta}$, parameterized by $\theta$, which maps the perceptual context $(q, \mathcal{O}_{\text{init}})$ to an optimal search plan $\mathcal{P}$:
\begin{equation}
\mathcal{P} \sim \pi_{\theta}(\cdot | q, \mathcal{O}_{\text{init}})
\end{equation}
where $\mathcal{P} = \{a_1, \dots, a_N\}$ is a sequence of parallel search actions (e.g., rewriting, filtering).

The objective is to find a policy $\pi^*$ that maximizes the expected utility of the final outcome given the environmental constraints:
\begin{equation}
\pi^* = \mathop{\arg\max}_{\pi} \mathbb{E}_{\mathcal{P} \sim \pi(\cdot|q, \mathcal{O}_{\text{init}})} \left[ \mathbb{E}_{\mathcal{R} \sim \mathcal{E}(\mathcal{P})} [U(\mathcal{R}, q)] \right]
\end{equation}
where $\mathcal{E}(\mathcal{P})$ denotes the execution of the plan in the environment $\mathcal{E}$ to obtain results $\mathcal{R}$, and $U(\cdot)$ is the utility function.

\begin{table*}[t]
\centering
\caption{EASP Diagnosis-Planning Matrix with illustrative examples.}
\label{tab:planning-examples}
\small
\renewcommand{\arraystretch}{1.2}
\setlength{\tabcolsep}{3.5pt}
\begin{tabular}{@{}l l l l@{}}
\toprule
\textbf{State} & \textbf{Diagnosis} & \textbf{Strategy} & \textbf{Transformation Flow} ($q \xrightarrow{\text{\textit{Perception in $(q, O_{init})$}} } \mathcal{P}$) \\
\midrule
\textbf{Effective} 
& \textit{State:} High relevance & Preservation & $q$: ``iPhone 17 Pro'' $\xrightarrow{\text{Top results highly relevant}}$ $\mathcal{P}$: \{``iPhone 17 Pro''\} \\
\midrule
\multirow{2}{*}
{\textbf{Recall Failure}}& \textit{Gap:} Query Noise & Sanitization & $q$: ``laptttop wth 32G'' $\xrightarrow{\text{Typo \& unit `32G' unmatched}}$ $\mathcal{P}$: \{``laptop 32GB RAM''\} \\
& \textit{State:} Inventory Void & No Refinement & $q$: ``brand X water'' $\xrightarrow{\text{Valid intent but 0 items found}}$ $\mathcal{P}$: HALT \\
\midrule
\multirow{5}{*}{\textbf{Precision Failure}} 
& \textit{Bias:} Entity drift & \multirow{5}{*}{Concretization} & $q$: ``bottoms match green shirt'' $\xrightarrow{\text{Results dominated by `shirts'}}$ $\mathcal{P}$: \{``khakis '', ``jeans''\} \\
& \textit{Bias:} Granularity mismatch & & $q$: ``camping'' $\xrightarrow{\text{Results cluttered w/ accessories}}$ $\mathcal{P}$: \{``tent'', ``sleeping bag''\} \\
& \textit{Gap:} Attribute misalignment & & $q$: ``high shrimp snack'' $\xrightarrow{\text{Items show shrimp ratio: 10\%, 60\%}}$ $\mathcal{P}$: \{``60\% ratio shrimp snack''\} \\
& \textit{Gap:} Scenario misalignment & & $q$: ``bird watching camera'' $\xrightarrow{\text{Mixed telephoto and wide-angle results}}$ $\mathcal{P}$: \{``telephoto camera''\} \\
\bottomrule
\multicolumn{4}{l}{\footnotesize \textit{Note:} Bias issues stem from retriever limitations; Gap issues stem from user-catalog vocabulary mismatch.} \\
\end{tabular}
\end{table*}

\subsection{Stage 1: Offline Data Synthesis via Teacher Agent}
\label{subsec:teacher_agent}
The Teacher Agent, based on a SOTA LLM, takes the original query $q$ and its retrieval snapshot $\mathcal{O}_{\text{init}}$ from the \textit{Retrieval Probe}, then analyzes the environment to diagnose retrieval failures and determine whether and how the query should be refined.

\noindent\textbf{Retrieval Probe.} 
Built upon the online search system, this module retains only core retrieval and relevance-matching capabilities, omitting computationally expensive components optimized for conversion. In our JD.com implementation, this design reduces tp99 latency by approximately 75\%, yet the retrieved items sufficiently reflect the underlying item distribution and expose the system's retrieval boundaries for diagnostic purposes.

\subsubsection{Perceptual Diagnosis}

The Teacher assesses the retrieval state by analyzing $\mathcal{O}_{\text{init}}$ in conjunction with $q$, categorizing it into one of three mutually exclusive states:

\begin{itemize}
    \item \textbf{Effective State:} $\mathcal{O}_{\text{init}}$ exhibits high relevance with sufficient coverage; no refinement is needed.

    \item \textbf{Recall Failure:} $\mathcal{O}_{\text{init}}$ contains few or zero results. The agent distinguishes two subcases:
    \begin{itemize}
        \item \textit{Query Noise:} The query contains elements that obstruct matching, including conversational fillers, marketing fluff, or lexical errors.
        \item \textit{Inventory Void:} The query expresses a valid intent, but no matching items exist in the catalog.
    \end{itemize}

    \item \textbf{Precision Failure:} $\mathcal{O}_{\text{init}}$ returns abundant results but with low precision. The agent further diagnoses the root cause along two dimensions:
    \begin{itemize}
        \item \textit{Bias Perception (Systematic Errors):} Problems arising from the retriever's inherent limitations, including \textbf{entity drift} (misinterpreting the head entity), \textbf{granularity mismatch} (retrieving peripheral items instead of core categories), and \textbf{negation failures}.
        
        \item \textit{Gap Perception (Semantic Misalignment):} Problems arising from the mismatch between user language and catalog vocabulary, including \textbf{attribute misalignment} (vague descriptors vs. concrete specifications) and \textbf{scenario misalignment} (usage scenarios vs. technical product attributes).
    \end{itemize}
\end{itemize}

\subsubsection{Adaptive Planning}

Guided by this diagnosis, Teacher Agent selects an optimized execution path (detailed in Table~\ref{tab:planning-examples}):

\begin{itemize}
    \item \textbf{Path A: Preservation (Effective State).} When retrieval is already effective, the agent preserves the original query to avoid over-optimization.

    \item \textbf{Path B: Sanitization (Recall Failure).} For recall failures caused by query noise, the agent performs \textbf{query sanitization}—removing non-informative elements and correcting lexical errors while preserving core intent. If an inventory void is detected, refinement is halted.

    \item \textbf{Path C: Concretization (Precision Failure).} For precision failures, the agent performs \textbf{intent concretization}, leveraging the LLM's reasoning and world knowledge to transform vague or misaligned queries into precise, retrievable forms:
    \begin{itemize}
        \item For \textit{bias} issues: re-anchoring drifted entities, focusing on core categories, or handling negated constraints.
        \item For \textit{gap} issues: mapping vague attributes to concrete specifications or translating usage scenarios to technical product attributes.
    \end{itemize}
\end{itemize}

\noindent\textbf{Diversity \& Validation.} 
To ensure the student planner learns a diverse policy space, we employ stochastic decoding during Teacher plan generation. For each $(q, \mathcal{O}_{\text{init}})$, we sample multiple trajectories with temperature $\tau > 0$, capturing the natural variability in valid reformulation strategies. All sampled plans are validated against the retrieval environment; only those yielding successful outcomes are retained. This yields a diverse set $\mathcal{D}_{\text{sft}}$ of execution-validated triples $(q, \mathcal{O}_{\text{init}}, \mathcal{P})$, exposing the Student to multiple viable behaviors for the same perceptual state.

\subsection{Stage 2: Planner Training and Business Alignment}

\label{subsec:planner_training}
\subsubsection{Policy Initialization via Supervised Fine-Tuning}
The Planner, based on a lightweight LLM, is fine-tuned on Teacher's diverse plan dataset $\mathcal{D}_{\text{sft}}$ using standard next-token prediction. This enables the model to internalize diagnostic reasoning patterns from offline demonstrations.

\subsubsection{Business Value Alignment via Reinforcement Learning}
\label{subsubsec:grpo}

We employ Group Relative Policy Optimization (GRPO)\cite{shao2024deepseekmath} to align the Planner with business outcomes.
For a given input context, the Planner samples $G$ plans $\{P_1, \dots, P_G\}$.

To compute the reward, each plan $P_i$ is executed against the online search engine to retrieve the Top-$K$ items, denoted as the set $\mathcal{D}_{P_i} = \{d_1, \dots, d_K\}$.
The reward $R(P_i)$ is defined as the average predicted conversion of these retrieved items, calculated using online production models:

\begin{equation}
    R(P_i) = \frac{1}{K} \sum_{d_j \in \mathcal{D}_{P_i}} \underbrace{\mathbb{I}(\phi_{\text{rel}}(q, d_j) \ge \tau)}_{\text{Hard Relevance Gate}} \cdot \underbrace{\phi_{\text{cvr}}(q, d_j)}_{\text{Predicted CVR}}
\end{equation}

Where:
\begin{itemize}
    \item $\phi_{\text{rel}}$ and $\phi_{\text{cvr}}$ are the existing Online Relevance and Conversion Prediction models, respectively.
    \item $\mathbb{I}(\cdot)$ is a binary indicator function acting as a \textit{Hard Relevance Gate}. If the raw relevance score is below the threshold $\tau$, the item's contribution is zeroed out (Reward=0). This strictly penalizes the planner for retrieving high-priced but irrelevant items, ensuring that conversion is grounded in semantic accuracy.
\end{itemize}

\subsection{Stage 3: Adaptive Online Serving via Dynamic Query Routing}
\label{subsec:online_serving}

Instead of uniformly processing all requests, we introduce a Complexity-Aware Router (an even smaller language model compared with Planner) to selectively activate the EASP pipeline:
\begin{itemize}
    \item \textbf{Fast Path:} Queries targeting specific products (e.g., "iPhone 17") bypass the planner entirely. This ensures nearly zero additional latency for the majority ($\sim$80\%) of traffic.
    \item \textbf{Complex Path:} Only complex queries trigger EASP.
\end{itemize}

\section{Experiments}
\label{sec:experiments}
\subsection{Experimental Setup}
\label{subsec:setup}

\textbf{Dataset Construction.}
We constructed our datasets from JD online search logs using a difficulty-aware strategy:

\begin{enumerate}
    \item \textbf{SFT Dataset}: 
    Real-world traffic is heavy-tailed (80\% simple, 20\% complex). We up-sampled complex queries and down-sampled simple queries (keeping 10\% for regularization) to build a 100k query sample corpus and instructed the Teacher Agent to generate initial SFT dataset. 
    \item \textbf{GRPO Subset}: 
    For the computationally expensive RL alignment, we selected top 5k queries targeting high-frequency queries with high reward standard deviations.
\end{enumerate}

\noindent \textbf{Implementation Details.} 
The Teacher Agent uses DeepSeek-R1\cite{guo2025deepseek}, and the Planner uses Qwen3-4B\cite{yang2025qwen3}. The query complexity router uses Qwen3-0.6B\cite{yang2025qwen3}. SFT was performed for 2 epochs. GRPO used a group size of $G=8$ on 8 NVIDIA H800 GPUs.

\subsection{Offline Evaluation}
\label{subsec:offline_eval}
\textbf{Test Set.} 
We constructed a test set of 10k complex queries sampled from historical logs. 

\noindent \textbf{Metrics.}
\begin{itemize}
    \item \textbf{REL@30}: Number of relevant items in Top-30 candidates.
    \item \textbf{HR@30}: Hitrate@30, which measures whether the finally purchased item appears in the top-30 retrieved candidates. To evaluate intent fulfillment, the ground truth includes purchases made through cross-channel interactions (e.g., recommendation feeds) or subsequent query reformulations within the same session, while maintaining the same underlying intent (same core product and brand).

\end{itemize}
\textbf{Baselines.} We compare EASP against 3 baselines:
\begin{itemize}

\item \textbf{Blind Rewriter}: 
An industry standard rewriter \cite{peng2024large,dai2024enhancing} using the same backbone (Qwen3-4B) and training (SFT+GRPO) as EASP, but conditioned solely on query $q$ without the retrieval snapshot $\mathcal{O}_{\text{init}}$, to ablate the environment-aware component.

    \item \textbf{w/o RL}: The same Qwen3-4B student trained on SFT dataset only. This isolates the impact of GRPO alignment.
    \item \textbf{ReAct Agent}: A standard multi-turn agent (based on DeepSeek-R1) with online tool access, representing the upper bound of reasoning, but with unacceptable latency in real online platform.
\end{itemize}
\textbf{Result Analysis.} 
As shown in Table \ref{tab:offline_results}, EASP achieves the optimal balance between performance and efficiency, outperforming baselines across key metrics while maintaining millisecond latency:

\textbf{Value of Environment Awareness.} Comparing EASP with the Blind Rewriter isolates the contribution of the Retrieval Probe. Removing probe access degrades all metrics: REL@30 drops from 23.3 to 20.7, HR@30 drops from 31\% to 28.6\%. These results validate that environment-aware planning is critical for retrieval effectiveness.

\textbf{Value of Business Alignment.} The w/o RL ablation achieves comparable REL@30 (23.0 vs. 23.3) to EASP, yet its HR@30 lags (29.5\% vs 31.0\%).This gap confirms that explicit alignment with conversion objectives via GRPO is essential for bridging relevance and business value.

\textbf{Comparison with ReAct.} While ReAct achieves the highest REL@30 (24.1), its gain over EASP is marginal (+0.8 pp). More importantly, ReAct's lower HR@30 (30.2\% vs. 31.0\%) reveals that general semantic relevance does not fully capture JD-specific conversion efficiency—a gap EASP bridges through business-aligned optimization. Crucially, ReAct incurs multi-second latency, whereas EASP maintains millisecond-level efficiency—the only solution delivering both high accuracy and real-time performance.

\begin{table}[h]
\centering
\caption{Offline Evaluation Result}
\label{tab:offline_results}
\small
\resizebox{\linewidth}{!}{
\begin{tabular}{l|c|c|c|c}
\toprule
\textbf{Method} & \textbf{REL@30} & \textbf{HR@30 (\%)} & \textbf{Latency} \\
\midrule
Blind Rewriter & 20.7 & 28.6 & low (ms level)\\
w/o RL  & 23.0 & 29.5 & low (ms level)\\
ReAct Agent & \textbf{24.1} & 30.2 & high (s level)\\
\midrule
\textbf{EASP(ours)} & 23.3 &\textbf{31.0} & low (ms level)\\
\bottomrule
\end{tabular}
}
\vspace{-0.2cm}
\end{table}

\subsection{Online A/B Testing}
\label{subsec:online_eval}

We conducted a two-week online A/B experiment on JD.com's AI-Search portal, exposing EASP to 10\% of live traffic. Compared to the production baseline (a Qwen3-4B planner trained with SFT and RL, but without environment probing), EASP achieved statistically significant gains.

On overall traffic (intent-to-treat), EASP lifted UCVR by \textbf{0.89\%} (p < 0.05) and GMV by \textbf{0.57\%} (p < 0.05). Among requests that actually triggered the new model (the complex path), UCVR increased by \textbf{4.10\%} and GMV by \textbf{2.59\%}. 

Regarding latency impact, the end-to-end planner latency reaches \textbf{20ms} at p75 for fast-path traffic and stays under \textbf{700ms} at p99 for complex queries, well within acceptable bounds given the conversion and revenue gains

Based on these results, EASP has been fully deployed in JD.com's AI-Search system.

\section{Conclusion and Future Work}
\label{sec:conclusion}

We propose EASP to resolve the "Blindness-Latency" dilemma in e-commerce search. The "Probe-then-Plan" mechanism grounds LLM reasoning in real-time retrieval snapshots, with GRPO alignment optimizing conversion and complexity-aware routing enabling efficient deployment. Online A/B testing on JD.com confirms significant UCVR and GMV lifts, demonstrating industrial viability.

Future work will extend EASP to incorporate user behavioral signals for personalized planning, adapting strategies based on individual preferences and interaction history.

\section*{Presenter Biography}
\label{bio}
\textbf{Mengxiang Chen} is a Senior Algorithm Engineer at JD.com, where he leads the development of agentic search systems and domain-specific LLMs for e-commerce. He holds a Master's degree from Shanghai Jiao Tong University. His research interests include LLM post-training, alignment, and intelligent information retrieval. 
\bibliographystyle{ACM-Reference-Format}
\bibliography{sample-base}

\end{document}